%% file: main.tex
\documentclass[runningheads]{llncs}

\usepackage{accv}

\usepackage{accvabbrv}

\usepackage{xcolor}
\usepackage{graphicx}
\usepackage{booktabs}

\usepackage{tabularray}

\usepackage{caption}
\usepackage{subcaption}

\usepackage{array}
\usepackage{algorithm}
\usepackage{algpseudocode}

\usepackage{multirow}
\usepackage{etoolbox}

\usepackage{wrapfig}
\usepackage{lipsum}

\usepackage{makecell}

\usepackage{colortbl}

\usepackage[accsupp]{axessibility}  

\usepackage{hyperref}

\usepackage{orcidlink}

\newcommand\freefootnote[1]{%
  \let\thefootnote\relax%
  \footnotetext{#1}%
  \let\thefootnote\svthefootnote%
}

\begin{document}

\title{InstantGeoAvatar:\\ Effective Geometry and Appearance Modeling of Animatable Avatars from Monocular Video} 

\titlerunning{InstantGeoAvatar: Animatable Avatars from Monocular Video}

\author{Alvaro Budria\inst{1}\orcidlink{0009-0000-9664-8829} \and
Adrian Lopez-Rodriguez\inst{2}\orcidlink{https://orcid.org/0000-0003-3984-5126} \and\\
Òscar Lorente*\inst{3} \and
Francesc Moreno-Noguer*\inst{4}\orcidlink{https://orcid.org/0000-0002-8640-684X}}

\authorrunning{A.~Budria et al.}

\institute{Institut de Robòtica i Informàtica Industrial (CSIC-UPC) \\ \email{alvaro.francesc.budria@upc.edu} \and
Vody \and
Floorfy \and
Amazon}

\maketitle

\begin{abstract}
We present InstantGeoAvatar, a method for efficient and effective learning from monocular video of detailed 3D geometry and appearance of animatable implicit human avatars. Our key observation is that the optimization of a hash grid encoding to represent a signed distance function (SDF) of the human subject is fraught with instabilities and bad local minima. We thus propose a principled geometry-aware SDF regularization scheme that seamlessly fits into the volume rendering pipeline and adds negligible computational overhead. Our regularization scheme significantly outperforms previous approaches for training SDFs on hash grids. We obtain competitive results in geometry reconstruction and novel view synthesis in as little as five minutes of training time, a significant reduction from the several hours required by previous work. InstantGeoAvatar represents a significant leap forward towards achieving interactive reconstruction of virtual avatars.

\end{abstract}

\freefootnote{*Work done while at Institut de Robòtica i Informàtica Industrial (CSIC-UPC).}

\freefootnote{The project website is at \href{https://github.com/alvaro-budria/InstantGeoAvatar}{github.com/alvaro-budria/InstantGeoAvatar}.}

\input{sections/01_introduction}

\input{sections/02_related_work}

\input{sections/03_method}

\input{sections/04_experiments}

\input{sections/05_limitations_future_work}


\section*{Acknowledgements}

This work has been supported by the project MOHUCO PID2020-120049RB-I00 funded by MCIU/AEI/10.13039/501100011033, and by the project GRAVATAR PID2023-151184OB-I00 funded by MCIU/AEI/10.13039/501100011033 and by ERDF, UE.

\clearpage  

%
%
\bibliographystyle{splncs04}
\bibliography{main}
\end{document}

%% file: sections/01_introduction.tex
\section{Introduction}
\label{sec:introduction}

Enabling the reconstruction and animation of 3D clothed avatars is a key step to unlock the potential of emerging technologies in fields such as augmented reality (AR), virtual reality (VR), 3D graphics and robotics. Interactivity and fast iteration over reconstructions and intermediate results can help speed up workflows for designers and practitioners. Different sensors are available for learning clothed avatars, including monocular RGB cameras, depth sensors and 4D scanners. RGB videos are the most widely available, yet provide the weakest supervisory signal, making this configuration elusive.

\begin{figure}[t!]
  \centering
  \includegraphics[width=0.95\linewidth]{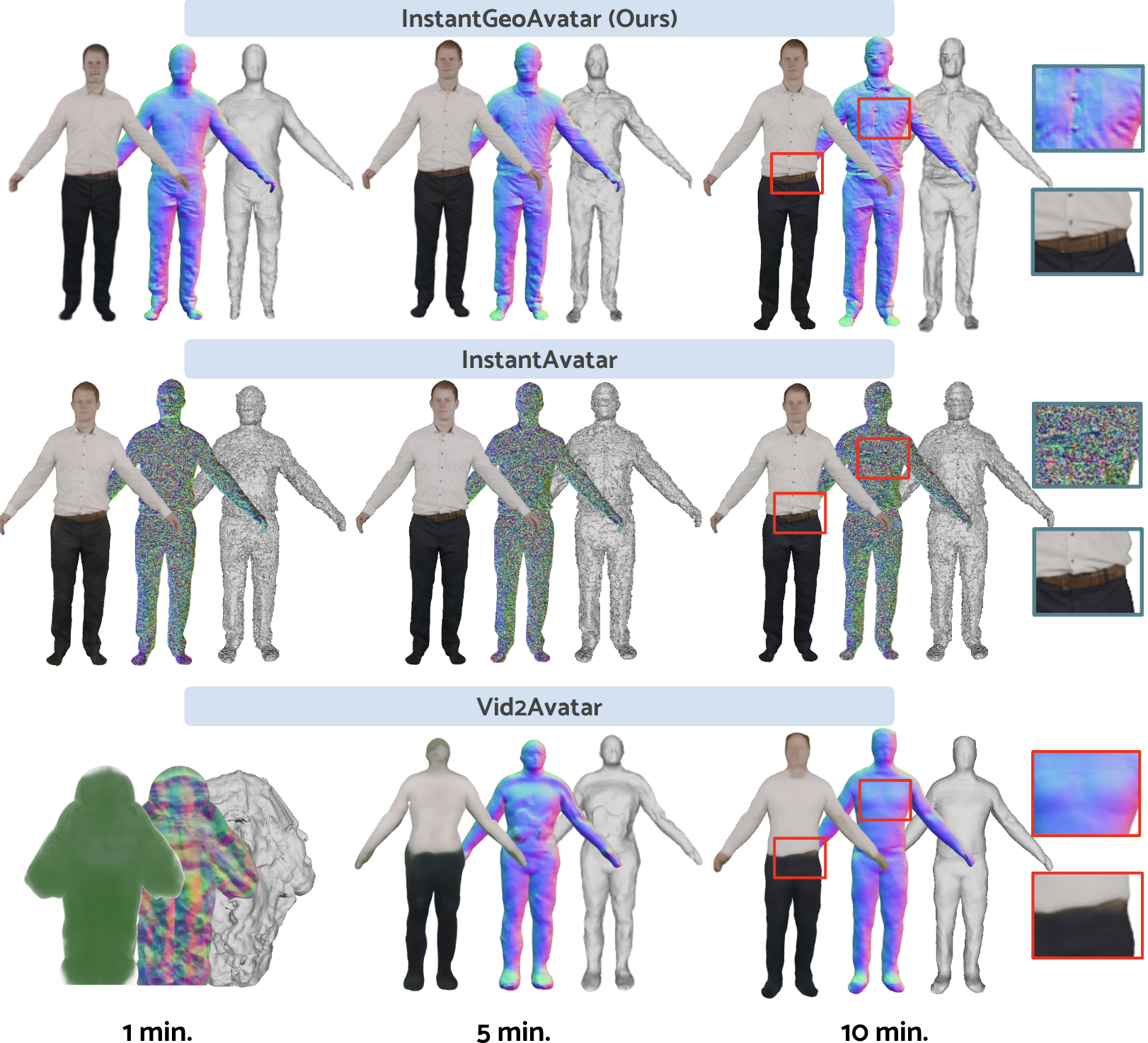}
  \caption{\textbf{InstantGeoAvatar.} We introduce a system capable of reconstructing the geometry and appearance of animatable human avatars from monocular video in less than 10 minutes.
  In order to attain high quality geometry reconstructions, we propose a smoothing term directly on the learned signed distance field during optimization, requiring no extra computation or sampling and delivering noticeable qualitative improvements.
  }
  \label{fig:training_progression}
\end{figure}

Traditional depth-based methods fusing depth measurements over time~\cite{depth_dynamicfusion, depth_kinectfusion, depth_articulation, depth_neuraldeform} produce compelling reconstructions but need complex and expensive setups to avoid sensor noise. Dense multi-view capture systems~\cite{MVS2, MVS1, MVS3, MVS4, MVS5, MVS6, MVS7, MVS8, MVS9, MVS10, MVS11, MVS12_shade, MVS13, MVS14} offer detailed 3D reconstructions by leveraging multi-view images and cues like silhouette or stereo. However, they can only be applied in controlled camera studios. Moreover, neither depth-based nor dense multi-view approaches can effectively work from RGB inputs alone or produce satisfactory results within short fitting times.

Mesh-based approaches~\cite{explicit1,explicit2,explicit3,explicit4,explicit5, explicit51, explicit6, explicit7, explicit8, explicit9, explicit10} struggle with garment deformations and are restricted to low-resolution topologies. Point cloud-based techniques~\cite{pointbased1, pointbased2, pointbased3, pointbased4, pointbased6, pointbased7, pointbased8, pointbased9, pointavatar, closet, implicitpointbased} have shown promising outcomes, but are not yet capable of fast-training times with RGB supervision only.

The advent of neural radiance fields (NeRFs)~\cite{nerf} enabled techniques for novel view synthesis and animation of human avatars from RGB image supervision only~\cite{instant_nvr, instantavatar, hnerf, anerf}. Volume rendering-based approaches typically learn a canonical representation that is deformed with linear blend skinning~\cite{SMPL} and an additional non-rigid deformation~\cite{anisdf, humanerf,monohuman}. Despite producing good renderings of human avatars, these techniques lack awareness of the underlying geometry. In parallel, some works have adopted a signed distance function (SDF)~\cite{volsdf} as basic primitive for learning clothed human avatars from 3D scans~\cite{pifu, phorhum, econ}. To remove the need for 3D supervision, some works have embedded SDFs within a volumetric rendering pipeline~\cite{vid2avatar, arah, anisdf}. Significant steps have been taken to speed up training of NeRF-based approaches~\cite{instant_nvr, instantavatar} by leveraging an efficient hash-grid spatial encoding~\cite{hashgridencoding}. Subsequent work has tried to improve training on such unstable and noisy hash grids~\cite{neuralangelo, shinobi} with some success.
Up to date, however, fast geometry learning in general and effective use of hash grid-based representations in particular for clothed human avatars with RGB supervision only remains elusive.

The challenges faced by NeRF- and SDF-based approaches trained with volume rendering can be succinctly reduced to effectively capturing realistic non-rigid deformations, dealing with noisy pose and camera estimates, slow training, and unstable training in the case of hash grid-based methods. In this paper, we specifically focus on the last two challenges, and aim at significantly advancing towards the realization of interactive use of human avatar modelling. We propose InstantGeoAvatar, a system capable of yielding good rendering and reconstruction quality in as little as 5 minutes of training, down from several hours as in prior work. Building on recent advances for fast training of NeRF based systems~\cite{hashgridencoding, instantavatar} and efficient training of hash grid encodings~\cite{neuralangelo, shinobi}, we demonstrate that even in combination such prior improvements and techniques are insufficient for fast and effective learning of 3D clothed humans. Thus we propose a simple yet effective regularization scheme that imposes a local geometric consistency prior during optimization, effectively removing undesired artifacts and defects on the surface.
The proposed approach, which effectively constrains surface curvature and torsion over continuous SDFs along ray directions, is easy to implement, fits neatly within the volume rendering pipeline, and delivers noticeable improvements over our base model without additional cost.

Our experiments demonstrate the effectiveness of the proposed method for effective and fast learning of animatable 3D human avatars from monocular video. At the short-training regime, InstantGeoAvatar yields superior geometry reconstruction and rendering quality compared to previous work in less than 10 minutes (see Fig.~\ref{fig:training_progression}).
While SoTA methods can yield more accurate reconstructions after several hours upon convergence, InstantGeoAvatar still shows comparable and even superior results with out of distribution (OOD) poses.
In addition, the presented ablation demonstrates that previous work on improving training of hash grid-based representations is insufficient for obtaining satisfying geometry reconstructions, highlighting the suitability of our proposal.

%% file: sections/02_related_work.tex
\section{Related Work}
\label{sec:relatedwork}

\subsection{Reconstructing Humans with Multi-View and Depth}

Traditional depth-based approaches for human shape reconstruction fuse depth measurements over time~\cite{depth_dynamicfusion, depth_kinectfusion}, and subsequent works improve robustness by incorporating additional priors~\cite{depth_articulation, depth_neuraldeform, depth_tracking, depth_parametric}, and by using custom-design sensors~\cite{depth_function4d, depth_sensors1}. Although effective, these methods require accurate depth information, which can only be obtained with laborious setups.
Methods based on calibrated dense multi-view capture setups~\cite{MVS2, MVS1, MVS3, MVS4, MVS5, MVS6, MVS7, MVS8, MVS9, MVS10, MVS11, MVS12_shade, MVS13, MVS14}  are capable of producing high-fidelity 3D reconstruction of human subjects by using multi-view images and other cues such as silhouette, stereo or shading. They are expensive and require skilled labor to operate, which hinders their applicability outside of camera studios. Both depth-based and dense multi-view cameras cannot work from monocular RGB inputs alone by design.

\subsection{Reconstructing Humans from Monocular \& Sparse Multi-views}

Template and mesh-based approaches can yield reasonable results even in the low-data regime by leveraging a low-rank human shape prior. On the other hand, implicit representations are continuous by design and have been used to produce detailed reconstructions of a clothed human body.
A specific subline of work aims at speeding up training of radiance field-based methods while maintaining good reconstruction quality. More recently, methods for human body modelling based on Gaussian Splats have shown compelling results at fast rendering times.

\noindent{\em Explicit Representations.}
Mesh-based techniques typically represent cloth deformations as deformation offsets~\cite{explicit1,explicit2,explicit3,explicit4,explicit5, explicit51, explicit6, explicit7, explicit8, explicit9, explicit10} added to a minimally clothed parametric human body model prior (e.g. SMPL~\cite{SMPL}) that has been previously rigged to a particular subject. Despite being highly compatible with parametric human body models, these approaches struggle with garment deformations and are limited to a low resolution topology.
Point cloud-based methods~\cite{pointbased1, pointbased2, pointbased3, pointbased4, pointbased6, pointbased7, pointbased8, pointbased9, pointavatar, closet, implicitpointbased} have shown promising results by combining the advantages of a representation that is explicit and simultaneously allows to model varying topologies.

\noindent{\em Neural Volumetric Rendering and Implicit Representations.}
Previous works have successfully applied neural radiance fields~\cite{nerf} and signed distance functions (SDFs)~\cite{volsdf,neus,unisurf} as basic primitives for modelling human avatars. Image based methods trained on 3D scans~\cite{geopifu, arch, archpp, pifu, pifuhd, icon, pamir, phorhum, surs, econ} and methods based on RGB-D and pointcloud data~\cite{explicit5, pina, explicit3, normalgan, occplanes, sfnet, rgbd1, rgbd2, difu, pc1, pc2} have difficulties with out-of-distribution poses, can fail to produce temporally consistent reconstructions, and suffer from incomplete observations in the case of RGB-D based methods. While some methods relying solely on 2D images attempt to train models that generalize to multiple subjects and can perform inference of novel views of a human directly
~\cite{nocanonic1, nocanonic2, nocanonic3}, most such methods involve volumetric rendering and learn a per-subject representation in a canonical space which can then be deformed to a given pose~\cite{humanerf, posemodulated, posevocab, avatarrex, monohuman, dualspacenerf, instant_nvr, anerf, hnerf, tava, anipixel, novelactor, humansoccluded, surfacealignednerf, dynamycmultiview, relightableanimatable, danbo, insta, anisdf, instantavatar, intrinsicavatar, arah, vid2avatar, xavatar, leap, snarf, fastsnarf, xavatar, invertibleskinning, neuralsurfacerecons, unif} or simply train a per-subject model without any canonical space~\cite{deliffas, rana, catnerf, pina, humanperformer}.

Significant effort has been devoted to improving the performance of volume rendering-based methods on various fronts. ~\cite{humanerf, vid2avatar, arah, neuman, anerf} jointly optimize neural network and body pose parameters. 
~\cite{xavatar, leap, snarf, fastsnarf, invertibleskinning, neuralsurfacerecons, unif, arah, surfacealignednerf} focus on the computation of correspondences between posed and canonical space. 
To improve the posing of the canonical representation, ~\cite{anisdf, monohuman, tava, snarf, fastsnarf,unif,tava,anipixel,novelactor} learn or refine skinning weights during training. ~\cite{humanerf, neuman, anisdf, arah, vid2avatar, hnerf, tava,anipixel,novelactor} add a non-rigid deformation module on top of the rigid deformation of a parametric body model.
~\cite{leap, monohuman,posevocab,posemodulated, danbo, anerf,novelactor} leverage pose encodings that help improve the results of pose-conditioned components.
~\cite{vid2avatar} allows segmenting a human on a video without mask supervision. Despite their success, these works do not demonstrate high geometry reconstruction quality at fast speeds.

An additional line of work aims at speeding up the fitting of models to each scene. Some works~\cite{instant_nvr, insta, instantavatar, intrinsicavatar, anipixel,humansoccluded} have employed a multiresolution hash grid spatial encoding~\cite{hashgridencoding}. Despite its effectiveness in accelerating training, it has been observed~\cite{neuralangelo, shinobi, baangp, rhino, smoothgradhash} that it lacks the implicit regularization towards lower frequencies that MLPs enjoy~\cite{spectralbias}. Even if the commonly used coordinate-based MLPs partially offset this bias~\cite{coordinatemlp} with a Fourier positional encoding, one can modulate the frequency bands of such encoding to still maintain a desirable level of smoothness, as done in common practice by works previously cited.
Some works have attempted to offset the lack of regularization in hash grids by considering the behaviour of gradients~\cite{neuralangelo, smoothgradhash}, implementing coarse-to-fine strategies~\cite{neuralangelo, baangp} and including a hybrid positional encoding to recover the regularizing effect~\cite{shinobi, rhino}. As shown in our experiments, these improvements are insufficient for obtaining quality reconstructions, hence the need for a more suitable training scheme, which we deliver in the form of an additional computational scheme in volume rendering.




\noindent{\em Gaussian Splats}~\cite{gsplatting0} have emerged as a powerful alternative primitive for volumetric rendering. Recent work has shown impressive results at the task of novel view synthesis of human avatars~\cite{gsplatting1, gsplatting2, gsplatting5, gsplatting6, gsplatting7, gsplatting8, gsplatting9, gsplatting10}, even achieving remarkably short training times~\cite{gsplatting_fast}, but we are unaware of any works specifically focusing on modeling the geometry of human avatars from monocular video.

%% file: sections/03_method.tex
\section{Method}
\label{sec:method}

InstantGeoAvatar learns a person-specific representation of geometry and appearance of a human subject from monocular video, given a set of input images with corresponding camera parameters, body masks and body poses.
We learn a parameterization of the canonical 3D geometry and the texture of a clothed human as an implicit signed distance field (SDF) and a texture field. A specialized canonicalization module finds rigid correspondences between posed and canonical spaces, and a non-rigid deformation module learns non-rigid clothing deformation and pose-dependent effects. The SDF and texture modules are learned via differentiable volume rendering, which is accelerated with an empty space skipping grid, as in previous work. 
We incorporate a surface regularization term within the volume rendering pipeline which not only ensures a nice surface smoothness and appearance, but also leads to watertight meshes, which other works struggle with.

\subsection{Preliminaries}
\label{sec:preliminaries}
We next describe the fundamental building blocks of the proposed pipeline, including an accelerated signed distance and texture field to model shape and appearance in canonical space and a canonicalization module combined with a non-rigid deformation to map this canonical representation into deformed space. 

\noindent{\bf Canonical Signed Distance Field.}
\label{sec:efficient-sdf}
We model human geometry in a canonical space with a signed distance function $\boldsymbol{f}_{sdf}$ that assigns a distance value and a feature vector to each point in the 3D canonical space:
\begin{align}
\boldsymbol{\textbf{f}}_{sdf}: \mathbb{R}^3 & \rightarrow \mathbb{R}, \mathbb{R}^{16} \\
x_c & \mapsto \textbf{d}, \textbf{v} \label{eq:sdf}
\end{align}

The body shape $\mathcal{S}$ in canonical space is then the zero-level set of $\boldsymbol{f}_{sdf}$:
\begin{align}
 \mathcal{S} = & \{ x_c\, |\, \boldsymbol{f}_{sdf} (x_c) = 0\}
\end{align}

Following~\cite{instantavatar} we use the multiresolution hash feature grid encoding from Instant-NGP~\cite{hashgridencoding} to parameterize $\boldsymbol{f}_{sdf}$.

\noindent{\bf Canonical Texture Field.} 
We learn a texture field $\boldsymbol{f}_{rgb}$ in canonical space that models the subject's appearance, conditioned on the SDF's predicted features:
\begin{align}
\boldsymbol{\textbf{f}}_{rgb}: \mathbb{R}^3, \mathbb{R}^{16} & \rightarrow \mathbb{R}^3 \label{eq:rgb}\\
x_c, \textbf{v} & \mapsto \textbf{c}
\end{align}

\noindent{\bf Articulating the Canonical Representation.}
We leverage the SMPL~\cite{SMPL} parametric body model to map a canonical point $x_c$ to a deformed point $x_d$ via linear blend skinning (LBS), according to a set of bone transformations $\boldsymbol{B}_i$ which are derived from body pose $\theta$:
\begin{align}
x_d = LBS(x) = \sum_{i=1}^{n_b} w_i \boldsymbol{B}_i x_c \label{eq:lbs}
\end{align}
 
However, the canonical correspondence $x_c^*$ of a deformed point $x_d$ is defined by the inverse mapping of Eq.~\ref{eq:lbs}. Thus it is necessary to compute the one-to-many mapping from points in posed space $x_d$ to their correspondences $x_c^*$ in canonical space. Fast-SNARF~\cite{fastsnarf} efficiently establishes such correspondences.

We add a pose-dependent offset to the rigid correspondence found by SNARF:
\begin{align}
\boldsymbol{f}_{\Delta x}: \mathbb{R}^3, \mathbb{R}^{69} & \rightarrow \mathbb{R}^3\\
x_c, \theta & \mapsto \Delta x,  \label{eq:nonrigid}
\end{align}


\subsection{Volume Rendering of SDF-based Radiance Fields}
\label{sec:rendering}

We learn the canonical representation end-to-end via differentiable volume rendering of our SDF function~\cite{volsdf}. As in previous works~\cite{arah, vid2avatar} we map the distance value to a density $\sigma$ by squashing it with the Laplacian Cumulative Distribution Function:
\begin{align}
\sigma(x_c) = & \alpha \Bigl( \frac{1}{2} + \frac{1}{2} sgn( - \boldsymbol{f}_{sdf} (x_c) ) (1 - \exp{- \frac{| \boldsymbol{f}_{sdf} | }{\beta}}) \Bigr), \label{eq:density} 
\end{align}
\noindent where $\beta$ is a learnable parameter and we set $\alpha = \frac{1}{\beta}$.

For a given pixel, we cast a ray $\textbf{r}(t) = \textbf{o} + t\textbf{l}$ from the optic center $\textbf{o}$ with ray direction $\textbf{l}$. We sample $N_p$ points between the near and far bounds in occupied space taking into account an occupancy grid, as presented in~\cite{instantavatar}. At each point, we query color $\boldsymbol{c}_i$ and density $\sigma_i$  from the canonical representation by warping ray points $\{ x^i_d \}^{N_p}$ from deformed to canonical space.

We then integrate the queried radiance and color values along each ray to get the rendered pixel color $\hat{C}$ as
\begin{align}
\hat{C} = \sum_{i=1}^{N_p} \alpha_i \prod_{i<j} (1-\alpha_j)\boldsymbol{c}_i,  \text{ with } \alpha_i = 1 - \exp(\sigma_i \delta_i),
\end{align}
\noindent where $\delta_i = || x_c^{i+1} - x_c^i || $ is the distance between consecutive samples.



\subsection{Training Objectives}
\label{sec:objectives}
We optimize our model against multiple weighted loss functions, including a smooth surface regularization term $\mathcal{L}_{smooth}$ that significantly boosts reconstruction quality:
\begin{align}
\mathcal{L} = \lambda_{rgb} \mathcal{L}_{rgb} + \lambda_{\alpha} \mathcal{L}_{\alpha} + \lambda_{Eik} \mathcal{L}_{Eik} + \lambda_{smooth} \mathcal{L}_{smooth}.
\label{eq:loss-rgb}
\end{align}

\noindent $\mathcal{L}_{rgb}$ is the photometric loss for rendered pixel color, $\mathcal{L}_{\alpha}$ is an L1 loss between the ground-truth and rendered masks and helps guide the reconstruction, and $\mathcal{L}_{Eik}$ corresponds to the Eikonal loss term~\cite{igr}, which encourages the learned SDF to have a well-behaved wave-like transition between consecutive isosurface slices. The loss weightings are $(\lambda_{rgb},~\lambda_{\alpha},~\lambda_{Eik},~\lambda_{smooth}) = (10,~0.1,~0.1,~1.0)$. $\mathcal{L}_{smooth}$ can be flexibly set within the range [0.5, 1.5] with good smoothing and details balance.

\begin{figure}[t!]
  \centering
  \includegraphics[width=.85\linewidth]{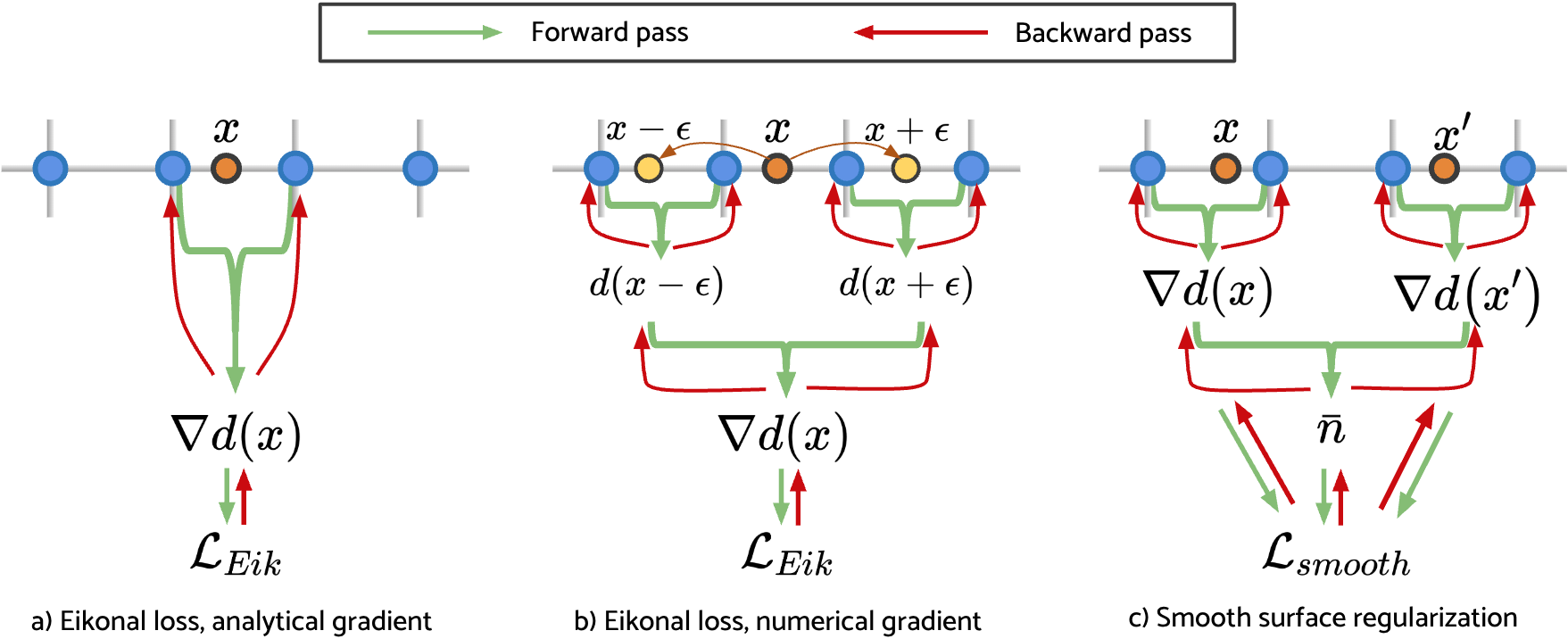}
    \caption{\textbf{Non-local updates of the hash grid features}. We consider a 1D hash grid encoding segment to illustrate how the proposed regularization affects backpropagation updates. Vanilla Eikonal loss (a) performs backpropagation updates on a single local hash grid cell resulting in discontinuous and spatially disconnected updates. (b)~\cite{neuralangelo}    used numerical gradients to distribute backpropagation updates to other cells in the grid, resulting in more spatially coherent learned features. Our proposed smooth surface regularization (c) also distributes backpropagation updates.}
  \label{fig:non-local-updates-hash}
\end{figure}

\noindent{\bf Surface Regularization.}
\label{sec:losses-smooth-surface-reg}
Directly substituting the NeRF~\cite{nerf} rendering module with the VolSDF~\cite{volsdf} rendering scheme explained in Sec.~\ref{sec:rendering} produces noisy, stripped surfaces (Fig.~\ref{fig:nosmooth}). 
The photometric loss only constrains the ray integral to render the desired color, but there are infinitely many shape variations that satisfy this condition. 
MLP-based architectures~\cite{arah, vid2avatar} are naturally biased to low-energy solutions~\cite{spectralbias}, but a hash-grid based representation lacks such implicit bias.

We devise a regularization procedure that effectively imposes a local coherence prior to the surface, by constraining surface normals at consecutive points on a ray to have the same direction and magnitude, i.e. that the isosurface be flat along the ray direction at each interval. 
If we locally approximate an arc parameterization of the surface with a straight line along the ray direction $\textbf{l}$ at ray depth $t$, this scheme amounts to taking the directional derivative $\frac{\textrm{d}\textbf{n}}{\textrm{d}t}$ of the normal of the surface $\textbf{n}$ along $\textbf{l}$ at $t$. 
By the Frenet–Serret formulas~\cite{frenet-serret} this term is related to the curvature $\kappa$ and torsion $\tau$ as

\begin{equation}
\frac{\textrm{d}\textbf{n}}{\textrm{d}t} = -\kappa \textbf{T} + \tau \textbf{B},\label{eq:dir-derivative}
\end{equation}
with $\textbf{T}$ the surface tangent vector and $\textbf{B}$ the surface binormal vector. Thus, by minimizing this term, we are directly imposing a penalty on the amount of curvature (how much the surface deviates from a straight line) and torsion (how much the surface deviates from following a regular path) that the surface presents along this direction, effectively enforcing the surface to be well-behaved.

We incorporate this regularization strategy into volume rendering without additional sampling by computing finite differences on the estimated normals $\nabla_x d(x_c)$ at each sampled point $x_c$ in canonical space, which we already obtained for the Eikonal loss term $\mathcal{L}_{Eik}$:
\begin{equation}
\mathcal{L}_{smooth} = \frac{1}{N_r} \sum_{i=1}^{N_r} \frac{1}{N^i_s} \sum_{j=1}^{N^i_s - 1} || \overline{\textbf{n}}_i^j - \hat{\textbf{n}}_i^j||_2 + || \overline{\textbf{n}}_i^j - \hat{\textbf{n}}_i^{j+1}||_2\label{eq:loss-smooth}
\end{equation}

\noindent where $\hat{\textbf{n}}_i^j = \frac{ \nabla_x d(x)}{|| \nabla_x d(x) ||_2}$ is the normalized estimated surface normal at point $x_i^j$,
$\overline{\textbf{n}}_i^j = \frac{1}{2} (\hat{\textbf{n}}_i^j + \hat{\textbf{n}}_i^{j+1})$ is the estimated normal at the midpoint of the interval $(x_i^j, x_i^{j+1})$, $N^i_s$ is the number of samples on the $i$-th ray and $N_r$ is the number of rays.
Moreover, it has been noted~\cite{neuralangelo} that surface normals $\nabla_x d(x)$ computed on hash grid-based representations only depend on the voxel features that immediately surround the point $x$. When learning with gradients from the Eikonal loss $\mathcal{L}_{Eik}$, this results in spatially discontinuous and incoherent updates that are limited to a single grid cell (Fig.~\ref{fig:non-local-updates-hash}a). Thus, by combining normals computed at different points with our approach (Fig.~\ref{fig:non-local-updates-hash}c), a greater number of adjacent and nearby grid cells are involved in the computations, and the resulting updates from the loss term are more stable spatially coherent.


\noindent{\bf Discussion.} Both on a conceptual and computational level, our proposed term is significantly different from the Eikonal loss~\cite{igr}, the use of finite differences for surface normal computation~\cite{neuralangelo}, and the curvature loss~\cite{neuralangelo} from previous works. The Eikonal loss $\mathcal{L}_{Eik}$ has a variational motivation and is derived as the solution of a wave propagation PDE. It acts as a global constraint on the SDF, ensuring it propagates through space as an onion, without silos or isolated components. Our constraint is geometric and differential as it works on local derivatives, and does not involve the magnitude of the gradient at a fixed point, but the differences of the normals along ray directions. 

Our approach is similar to that of~\cite{neuralangelo} in that we also propagate gradient updates to multiple cells for more stable training. But unlike them, we achieve this effect by considering the surface normals at consecutive ray points, and use more precise analytical gradients to compute those normals. We only take finite differences along ray directions to compute the derivative of the surface normal.

The curvature loss from~\cite{neuralangelo} computes the Laplacian considering how much an SDF value at each sample point diverges from the SDF values of neighbors at a fixed distance, which in three-dimensional space is not explicitly related to the curvature and torsion of the surface. Moreover, computing it involves 6 additional samples and relies on a fixed scale on the finite differences which needs to be scheduled over training. On the other hand, our approach explicitly relates the directional derivative of the surface normals to geometric quantities of curvature and torsion. Since the rays' direction of incidence on the surface varies with pixel location and camera location, the regularization is effectively applied multi-scale over training (Fig.~\ref{fig:multiscale}), without the need for any scale scheduling nor any additional samples.

\label{sec:objectives}

%% file: sections/04_experiments.tex
\section{Experiments}
\label{sec:experiments}

We evaluate the rendering and geometry reconstruction quality of our proposed method and compare it with other SoTA methods both on in-distribution and out-of-distribution poses. Additionally, we present an ablation that investigates the effect of different techniques proposed for stable training with hash-grid encodings and validate the effectiveness of our approach. Further results are presented in the supplementary video.

\subsection{Datasets}

\noindent{\bf{PeopleSnapshot~\cite{explicit10}}}
contains videos of humans rotating in front of a camera, and is the main dataset employed by InstantAvatar~\cite{instantavatar}. For a fair comparison, we follow its same evaluation protocol,   taking the pose parameters optimized by Anim-NeRF\cite{animnerf} and keeping them frozen throughout training.

\noindent{\bf{X-Humans~\cite{xavatar}}}  provides several video sequences of humans in various poses and performing different actions. The dataset contains accurate ground-truth detailed clothed 3D meshes and   SMPL fits for each subject at each frame. 
We select a subset of 5 sequences to perform our experiments. Each sequence contains several tracks or subsequences. For each sequence we pick a set of tracks amounting to between 150 to 300 consecutive frames for training and another set of tracks for evaluation, containing both in-distribution as well as out-of-distribution poses that significantly depart from those poses seen during training.

\begin{table}[t!]
\scriptsize
 \renewcommand{\tabcolsep}{2.5pt}
 \begin{minipage}[t]{.48\linewidth}
 \caption{\textbf{Geometry Reconstruction}. We report L2 Chamfer Distance (CD), Normal Consistency (NC) and Intersection over Union (IoU) on the X-Humans dataset~\cite{xavatar}.}
 \label{tab:reconstruction_in}
 \centering
  \renewcommand{\arraystretch}{1.1}
 \begin{tabular}{ c  c  c  c  c  c}
 Sequence & Metric & V2A & V2A\textsuperscript{10} & IA & Ours \\\hline

 \multirow{3}{*}{25} & CD $\downarrow$ & \textbf{0.304} & 0.602 & 0.901 & \underline{0.579} \\
 & NC $\uparrow$ & \textbf{0.919} & 0.894 & 0.688 & \underline{0.910} \\
 & IoU $\uparrow$ & \textbf{0.977} & 0.920 & 0.842 & \underline{0.974} \\\hline
 
 \multirow{3}{*}{28} & CD $\downarrow$ & \textbf{0.393} & 0.842 & 0.851 & \underline{0.600} \\
 & NC $\uparrow$ & \textbf{0.922} & 0.871 & 0.727 & \underline{0.914} \\
 & IoU $\uparrow$ & \textbf{0.963} & 0.902 & 0.755 & \underline{0.947} \\\hline
 
 \multirow{3}{*}{35} & CD $\downarrow$ & \textbf{0.417} & 0.641 & 1.073 & \underline{0.557} \\
 & NC $\uparrow$ & \textbf{0.912} & 0.879 & 0.673 & \underline{0.891} \\
 & IoU $\uparrow$ & \textbf{0.944} & 0.902 & 0.798 & \underline{0.923} \\\hline
 
 \multirow{3}{*}{36} & CD $\downarrow$ & \textbf{0.572} & 0.748 & 1.494 & \underline{0.691} \\
 & NC $\uparrow$ & \textbf{0.844} & 0.817 & 0.576 & \underline{0.838} \\
 & IoU $\uparrow$ & \textbf{0.950} & 0.915 & 0.771 & \underline{0.936} \\\hline
 
 \multirow{3}{*}{58} & CD $\downarrow$ & \textbf{0.311} & \underline{0.512} & 1.693 & 0.597 \\
 & NC $\uparrow$ & \textbf{0.930} & \underline{0.887} & 0.681 & 0.875 \\
 & IoU $\uparrow$ & \textbf{0.973} & \underline{0.964} & 0.704 & 0.950 \\\hline

 \end{tabular}
 \end{minipage} \quad
 \begin{minipage}[t]{.48\linewidth}
 \caption{\textbf{Geometry Reconstruction}. We report L2 Chamfer Distance (CD), Normal Consistency (NC) and Intersection over Union (IoU) on OOD poses of the X-Humans dataset~\cite{xavatar}.}
 \label{tab:reconstruction_out}
 \centering
  \renewcommand{\arraystretch}{1.1}
 \begin{tabular}{ c  c  c  c  c  c}
 Sequence & Metric & V2A & V2A\textsuperscript{10} & IA & Ours \\\hline
 \multirow{3}{*}{25} & CD $\downarrow$ & \textbf{0.751} & 0.892 & 1.72 & \underline{0.881} \\
 & NC $\uparrow$ & \textbf{0.889} & 0.871 & 0.699 & \underline{0.879} \\
 & IoU $\uparrow$ & \textbf{0.931} & 0.928 & 0.845 & \underline{0.930} \\\hline
 
 \multirow{3}{*}{28} & CD $\downarrow$ &\underline{1.096} & 1.445 & 2.452 & \textbf{1.089} \\
 & NC $\uparrow$ & \underline{0.860} & 0.842 & 0.720 & \textbf{0.885} \\
 & IoU $\uparrow$ & \textbf{0.911} & 0.901 & 0.761 & \underline{0.909} \\\hline
 
 \multirow{3}{*}{35} & CD $\downarrow$ & \textbf{0.705} & 0.804 & 1.213 & \underline{0.800} \\
 & NC $\uparrow$ & \textbf{0.883} & 0.869 & 0.678 & \underline{0.871} \\
 & IoU $\uparrow$ & \textbf{0.945} & \underline{0.912} & 0.794 & 0.896 \\\hline
 
 \multirow{3}{*}{36} & CD $\downarrow$ & \textbf{0.760} & 0.903 & 2.23 & \underline{0.871} \\
 & NC $\uparrow$ & \underline{0.842} & 0.830 & 0.579 & \textbf{0.845} \\
 & IoU $\uparrow$ & \underline{0.910} & 0.899 & 0.768 & \textbf{0.922} \\\hline
 
 \multirow{3}{*}{58} & CD $\downarrow$ & \textbf{0.732} & \underline{0.785} & 1.892 & 0.842 \\
 & NC $\uparrow$ & \underline{0.835} & 0.824 & 0.682 & \textbf{0.838} \\
 & IoU $\uparrow$ & \textbf{0.898} & 0.863 & 0.705 & \underline{0.877} \\\hline
 \end{tabular}
 \end{minipage}
\end{table}

\subsection{Baselines}

\noindent{\bf{InstantAvatar (IA)~\cite{instantavatar}}} 
  is capable of modeling the appearance of a human subject in short training times. In our experiments, we use the publicly available code to train this baseline for 10 minutes. Despite its ability to produce high-quality renderings, IA lacks geometric awareness as it models radiance.

\noindent{\bf{Vid2Avatar (V2A)~\cite{vid2avatar}}}  is able to produce high-quality geometry reconstructions without the need for ground-truth masks but requires several hours of training to reach convergence. We compare our method against Vid2Avatar after 24 hours (V2A) and 10 minutes (V2A\textsuperscript{10}) of training, to show that our method can achieve good rendering and reconstruction quality much faster.

\begin{table}[t!]
\scriptsize
 \renewcommand{\tabcolsep}{2.5pt}
 \begin{minipage}[t]{.48\linewidth}
 \caption{\textbf{Rendering Quality}. We report PSNR, SSIM and LPIPS on in-distribution poses of the X-Humans dataset~\cite{xavatar}.}
 \label{tab:rendering_in}
 \centering
  \renewcommand{\arraystretch}{1.1}
 \begin{tabular}{ c  c  c  c  c  c}
 Sequence & Metric & V2A & V2A\textsuperscript{10} & IA & Ours \\\hline

 \multirow{3}{*}{25} & PSNR $\uparrow$ & 29.82 & 25.12 & \textbf{30.05} & \underline{29.87} \\
 & SSIM $\uparrow$ & \textbf{0.991} & 0.960 & \underline{0.986} & 0.983 \\
 & LPIPS $\downarrow$ & 0.016 & 0.033 & \textbf{0.012} & 0.020 \\\arrayrulecolor{gray}\hline
 
 \multirow{3}{*}{28} & PSNR $\uparrow$ & \textbf{28.20} & 24.35 & 27.63 & \underline{27.74} \\
 & SSIM $\uparrow$ & \underline{0.980} & 0.952 & \textbf{0.982} & 0.976 \\
 & LPIPS $\downarrow$ & \textbf{0.017} & 0.025 & \underline{0.019} & 0.032 \\\hline
 
 \multirow{3}{*}{35} & PSNR $\uparrow$ & \underline{29.01} & 25.49 & \textbf{29.47} & 28.68 \\
 & SSIM $\uparrow$ & \underline{0.984} & 0.961 & \textbf{0.988} & 0.979 \\
 & LPIPS $\downarrow$ & \underline{0.019} & 0.032 & \textbf{0.011} & 0.028 \\\hline
 
 \multirow{3}{*}{36} & PSNR $\uparrow$ & 29.59 & 26.77 & \textbf{31.74} & \underline{30.97} \\
 & SSIM $\uparrow$ & 0.973 & 0.957 & \textbf{0.981} & \underline{0.974}\\
 & LPIPS $\downarrow$ & \underline{0.015} & 0.030 & \textbf{0.010} & 0.021 \\\hline
 
 \multirow{3}{*}{58} & PSNR $\uparrow$ & 30.32 & 26.93 & \textbf{31.05} & \underline{30.54} \\
 & SSIM $\uparrow$ & \underline{0.986} & 0.959 & \textbf{0.989} & 0.985 \\
 & LPIPS $\downarrow$ & 0.018 & 0.026 & \textbf{0.009} & \underline{0.016} \\\arrayrulecolor{black}\hline
 \end{tabular}
 \end{minipage}  \quad 
 \begin{minipage}[t]{.48\linewidth}
 \caption{\textbf{Rendering Quality}. We report PSNR, SSIM and LPIPS on out-of-distribution poses of the X-Humans dataset~\cite{xavatar}.}
 \label{tab:rendering_out}
 \centering
  \renewcommand{\arraystretch}{1.1}
 \begin{tabular}{ c  c  c  c  c  c}
 Sequence & Metric & V2A & V2A\textsuperscript{10} & IA & Ours \\\hline
 \multirow{3}{*}{25} & PSNR $\uparrow$ & 23.09 & 19.66 & \underline{23.32} & \textbf{23.56} \\
 & SSIM $\uparrow$ & \textbf{0.965} & 0.937 & 0.957 & \underline{0.964} \\
 & LPIPS $\downarrow$ & \textbf{0.025} & 0.045 & 0.032 & \underline{0.027} \\\arrayrulecolor{gray}\hline
 
 \multirow{3}{*}{28} & PSNR $\uparrow$ &24.50 & 20.93 & \underline{24.56} & \textbf{24.77} \\
 & SSIM $\uparrow$ & 0.963 & 0.942 & \underline{0.964} & \textbf{0.966} \\
 & LPIPS $\downarrow$ & \textbf{0.028} & 0.038 & \underline{0.030} & 0.033 \\\hline
 
 \multirow{3}{*}{35} & PSNR $\uparrow$ & \textbf{25.68} & 21.34 & \underline{25.23} & 25.21 \\
 & SSIM $\uparrow$ & \textbf{0.960} & 0.952 & 0.955 & \underline{0.958} \\
 & LPIPS $\downarrow$ & \underline{0.031} & 0.048 & \textbf{0.027} & 0.039 \\\hline
 
 \multirow{3}{*}{36} & PSNR $\uparrow$ & \textbf{26.41} & 22.05 & 25.07 & \underline{26.33} \\
 & SSIM $\uparrow$ & \underline{0.953} & 0.947 & 0.950 & \textbf{0.955} \\
 & LPIPS $\downarrow$ & 0.036 & 0.040 & \textbf{0.028} & \underline{0.031} \\\hline
 
 \multirow{3}{*}{58} & PSNR $\uparrow$ & \underline{23.88} & 21.40 & 23.82 & \textbf{24.05} \\
 & SSIM $\uparrow$ & \textbf{0.964} & 0.951 & 0.957 & \underline{0.960} \\
 & LPIPS $\downarrow$ & \textbf{0.025} & 0.040 & 0.033 & \underline{0.029} \\\arrayrulecolor{black}\hline
 \end{tabular}
 \end{minipage}
\end{table}

\begin{table}[t!]
 \scriptsize
 \renewcommand{\tabcolsep}{2.5pt}
 \begin{minipage}{.48\linewidth}
 \caption{\textbf{Rendering Quality}. We report PSNR, SSIM and LPIPS on in-distribution poses of the PeopleSnapshot dataset~\cite{explicit10}. Note that unlike in~\cite{instantavatar}, we do not overfit the poses prior to testing.}
 \label{tab:peoplesnapshot}
 \centering
  \renewcommand{\arraystretch}{1.1}
 \begin{tabular}{ c  c  c  c  c  c}
 Sequence & Metric & V2A & V2A\textsuperscript{10} & IA & Ours \\\hline

 \multirow{3}{*}{f-3-c} & PSNR $\uparrow$ & \textbf{24.96} & 22.91 & 24.24 & \underline{24.89} \\
 & SSIM $\uparrow$ & \textbf{0.957} & 0.916 & 0.949 & \underline{0.952} \\
 & LPIPS $\downarrow$ & \textbf{0.030} & 0.062 & \underline{0.033} & 0.046 \\\arrayrulecolor{gray}\hline
 
 \multirow{3}{*}{m-3-c} & PSNR $\uparrow$ &\textbf{28.37} & 20.93 & 27.96 & \underline{28.00} \\
 & SSIM $\uparrow$ & \textbf{0.963} & 0.942 & \textbf{0.963} & \underline{0.961} \\
 & LPIPS $\downarrow$ & \textbf{0.017} & 0.038 & 0.021 & 0.035 \\\hline
 
 \multirow{3}{*}{f-4-c} & PSNR $\uparrow$ & \textbf{27.44} & 21.34 & 26.89 & \underline{27.16} \\
 & SSIM $\uparrow$ & \textbf{0.968} & 0.952 & 0.959 & \underline{0.962} \\
 & LPIPS $\downarrow$ & \underline{0.026} & 0.048 & \textbf{0.023} & 0.040 \\\hline
 
 \multirow{3}{*}{m-4-c} & PSNR $\uparrow$ & \textbf{26.53} & 22.05 & 25.38 & \underline{25.46} \\
 & SSIM $\uparrow$ & \textbf{0.955} & 0.947 & \underline{0.953} & 0.951 \\
 & LPIPS $\downarrow$ & \textbf{0.039} & \underline{0.040} & 0.042 & 0.059 \\\arrayrulecolor{black}\hline
 
 \end{tabular}
 \end{minipage}\hfill
	\begin{minipage}{0.45\linewidth}
		\centering
		\includegraphics[width=52mm]{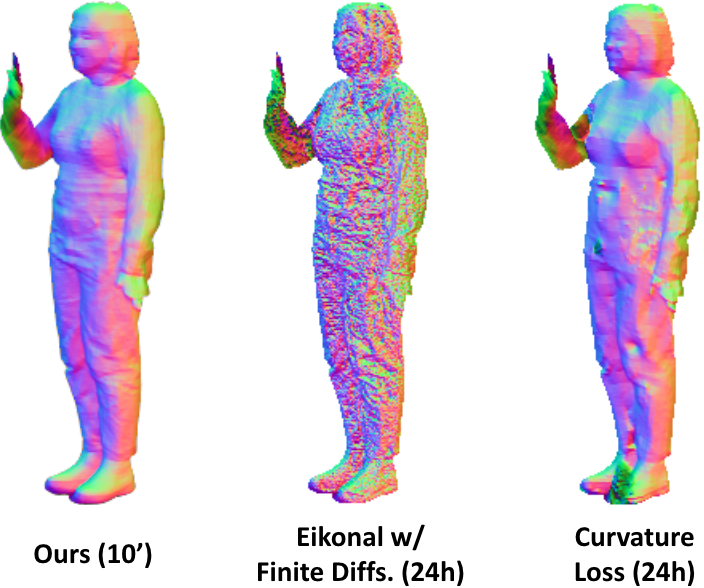}
		\captionof{figure}{\textbf{Neuralangelo's~\cite{neuralangelo} SDF training scheme at longer training regime.} Our approach beats Neuralangelo's proposal even after 24 hours of training.}
		\label{fig:ablation-24h}
	\end{minipage}
\end{table}

\subsection{Geometry Reconstruction Quality}

First, we compare our proposed human shape reconstruction approach to the baselines (Table~\ref{tab:reconstruction_in}). IA~\cite{instantavatar} fails to model meaningful surfaces. V2A can very accurately reconstruct human shapes, but with slow training times. V2A\textsuperscript{10} is initialized with a SMPL body shape, which gives it a significant boost in geometric accuracy at the beginning of training. However, as seen qualitatively in Figures~\ref{fig:training_progression} and~\ref{fig:qualitative_xhumans} it fails to model clothing details. Our method strikes a good balance between reconstruction accuracy and speed, obtaining reasonably good results within 10 minutes of training.

\begin{table}[!t]
 \scriptsize
 \renewcommand{\tabcolsep}{1.pt}
	\begin{minipage}{.48\linewidth}
 \caption{\textbf{Quantitative comparison of SDF regularization schemes}. We report PSNR, SSIM, Chamfer Distance (CD) and Normal Consistency (NC) on the X-Humans dataset~\cite{xavatar}.}
 \label{tab:ablation}
 \centering
  \renewcommand{\arraystretch}{2.75}
 \begin{tabular}{ c  c  c  c  c }
 Component & PSNR $\uparrow$ & SSIM $\uparrow$ & CD $\downarrow$ & NC $\uparrow$ \\[-0.1cm]\hline

 Base & 28.90 & 0.961 & 0.827 & 0.827 \\\arrayrulecolor{gray}\hline
 
 \makecell{ Eikonal w/ \\Finite Diffs.~\cite{neuralangelo}} & 28.60 & 0.960 & 2.45 & 0.657 \\\arrayrulecolor{gray}\hline
 
 \makecell{ Curvature\\Loss~\cite{neuralangelo}} & 28.51 & 0.957 & 2.03 & 0.536 \\\arrayrulecolor{gray}\hline

 \makecell{ Eik. w/ F. Diffs. \\and Curv. Loss~\cite{neuralangelo}} & 28.47 & 0.959 & 5.03 & 0.52 \\\arrayrulecolor{gray}\hline
 
 \makecell{Hybrid Pos. \\Encoding~\cite{shinobi}} & 26.48 & 0.954 & \textbf{0.752} & 0.803 \\\arrayrulecolor{gray}\hline
 
 $\mathcal{L}_{smooth}$ (ours) & \textbf{29.24} & \textbf{0.964} & 0.772 & \textbf{0.850} \\\arrayrulecolor{black}\hline

 \end{tabular}
 \end{minipage}\hfill
	\begin{minipage}{0.45\linewidth}
		\centering
		\includegraphics[width=45.5mm]{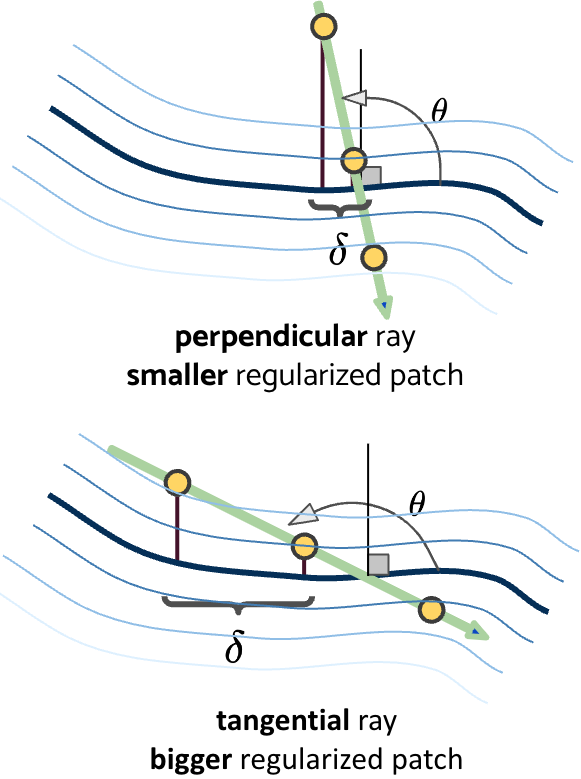}
		\captionof{figure}{\textbf{Multiscale effect of the proposed loss term.}}
		\label{fig:multiscale}
	\end{minipage}
\end{table}

\subsection{View Synthesis Quality}
\begin{figure}[b!]
  \centering
  \includegraphics[width=1.0\linewidth]{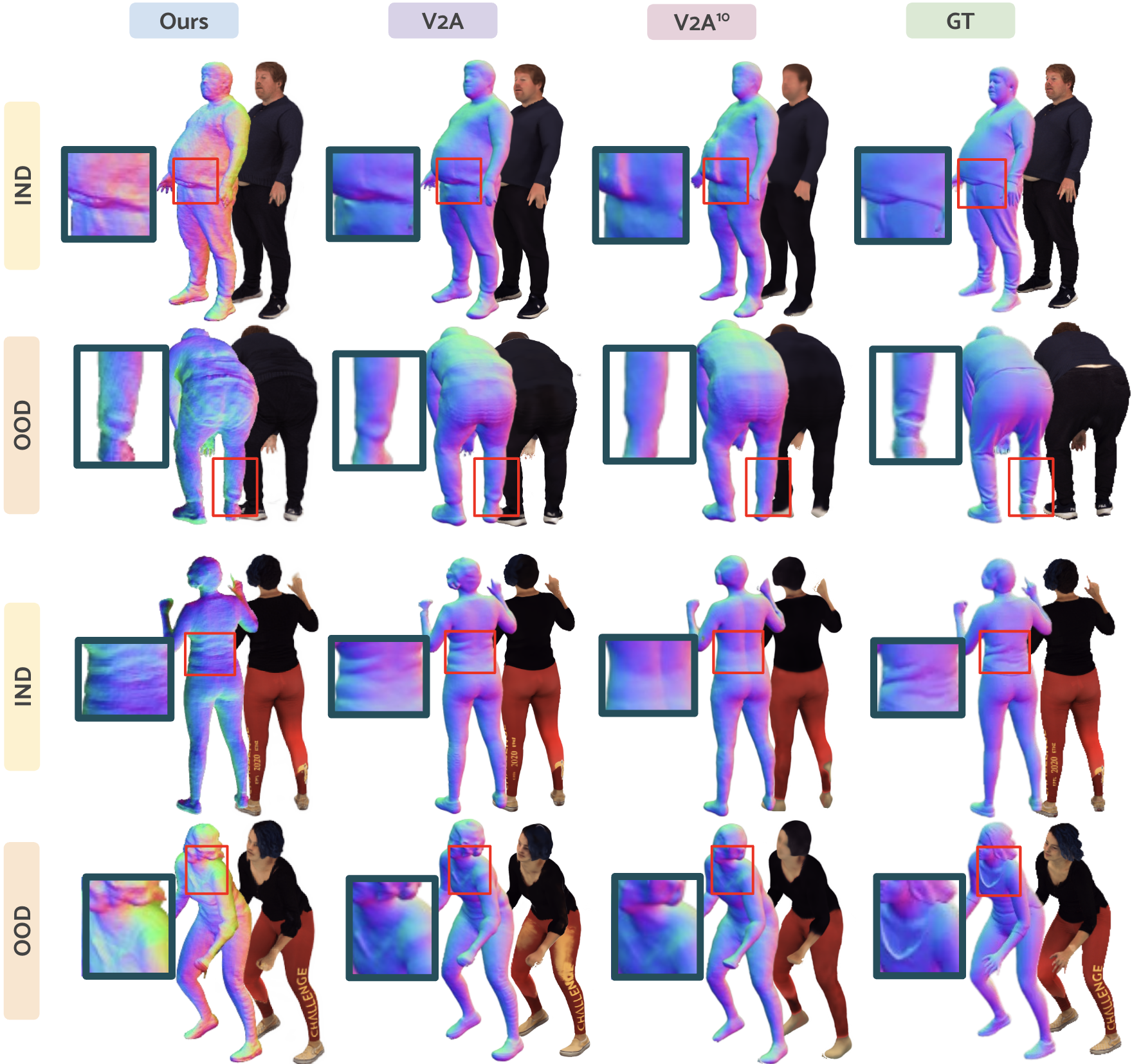}
  \caption{\textbf{Qualitative Results on Rendering and Reconstruction.} V2A produces very accurate and smooth reconstructions after several hours of training, while V2A\textsuperscript{10} fails to represent higher frequency details like the letters on the second subject's leg. InstantGeoAvatar can produce satisfying results in less than ten minutes.
  }
  \label{fig:qualitative_xhumans}
\end{figure}

Tables~\ref{tab:rendering_in} and \ref{tab:peoplesnapshot} show our method produces quality renderings on both datasets. Since Vid2Avatar is not designed for fast training, V2A\textsuperscript{10} fails to synthesize good renders, while both IA and InstantGeoAvatar achieve the same rendering quality as V2A in a much shorter time. As seen qualitatively in Figures~\ref{fig:training_progression} and~\ref{fig:qualitative_xhumans}, V2A\textsuperscript{10} yields flat-textured renders, whereas IA and InstantGeoAvatar can model texture patterns such as shirt wrinkles.

\begin{table}[!t]
 \renewcommand{\tabcolsep}{4.5pt}
 \renewcommand{\arraystretch}{1.2}
 \caption{\textbf{Computational demand of SDF regularization schemes}.}
\label{tab:ablation_speed}
 \centering
\begin{tabular}{cccccc}
 &
  Base &
  Ours &
  \begin{tabular}[c]{@{}c@{}}Hybrid Pos.\\ Encoding~\cite{shinobi}\end{tabular} &
  \begin{tabular}[c]{@{}c@{}}Eikonal w/\\ Finite Diffs.~\cite{neuralangelo}\end{tabular} &
  \begin{tabular}[c]{@{}c@{}}Curvature\\ Loss~\cite{neuralangelo}\end{tabular} \\ \hline training speed &
  \textbf{8.0 it./s} &
  \underline{7.7} it./s &
  4.1 it./s &
  5.3 it./s &
  4.8 it./s \\\arrayrulecolor{gray}\hline
peak memory &
  $\sim$ \textbf{15 GB} &
  $\sim$ \underline{15 GB} &
  $\sim$ 16 GB &
  $\sim$ 20 GB &
  $\sim$ 20 GB \\\arrayrulecolor{black}\hline
\end{tabular}
\end{table}

\begin{figure}[!t]
  \centering
  \includegraphics[width=0.85\linewidth]{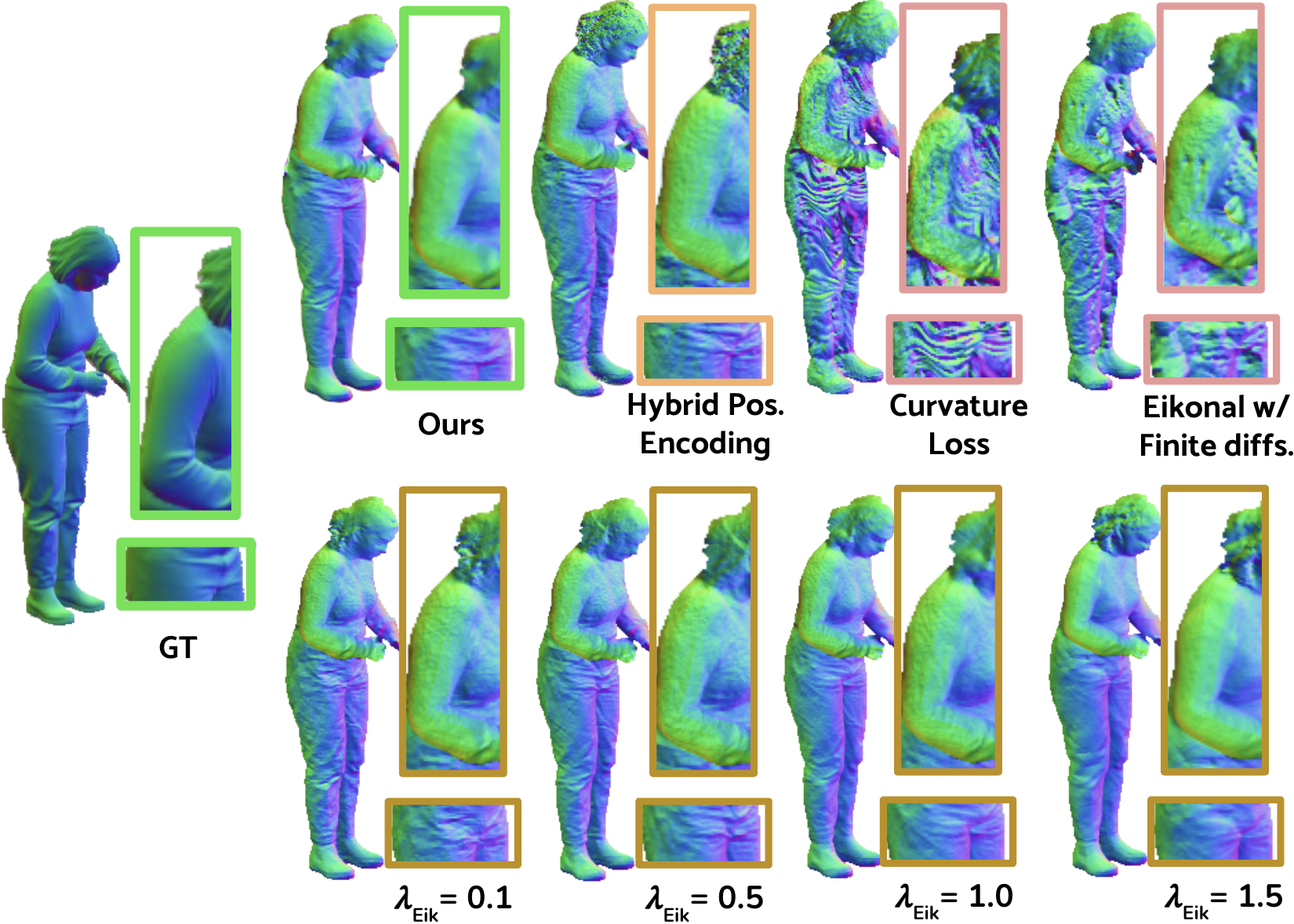}
  \caption{\textbf{Qualitative comparison of SDF regularization schemes.} From left and right, top to bottom: ours, hybrid positional encoding~\cite{shinobi}, curvature loss and finite differences derivatives~\cite{neuralangelo}, and varying weights for Eikonal loss~\cite{igr}.}
  \label{fig:comparison_methods}
\end{figure}

\subsection{Synthesis \& Reconstruction Quality on OOD Poses}
As seen in Table~\ref{tab:reconstruction_out}, the geometry produced by V2A degrades significantly presumably because it conditions its SDF representation on pose parameters. Our method performs almost identically as V2A, and outperforms InstantAvatar and V2A\textsuperscript{10}. This can also be observed in Fig.~\ref{fig:qualitative_xhumans} and in the supplementary video.

\subsection{Ablation Study}
\label{ablation}
\begin{figure}[t!]
  \centering
  \begin{subfigure}{0.49\linewidth}
    \includegraphics[width=1.\linewidth]{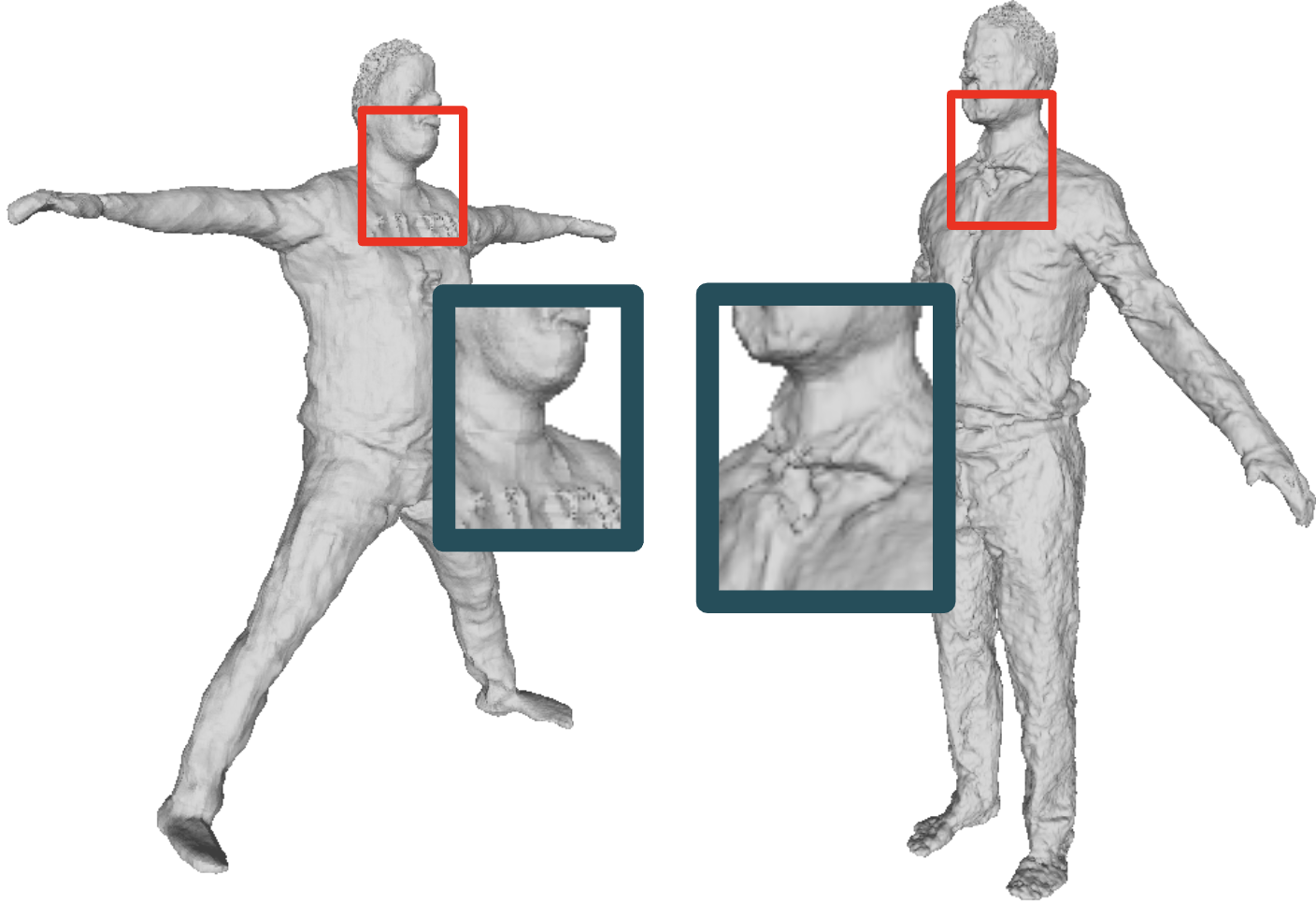}
    \caption{Without smoothing term.}
    \label{fig:nosmooth}
  \end{subfigure}
  \hfill
  \begin{subfigure}{0.49\linewidth}
    \includegraphics[width=1.\linewidth]{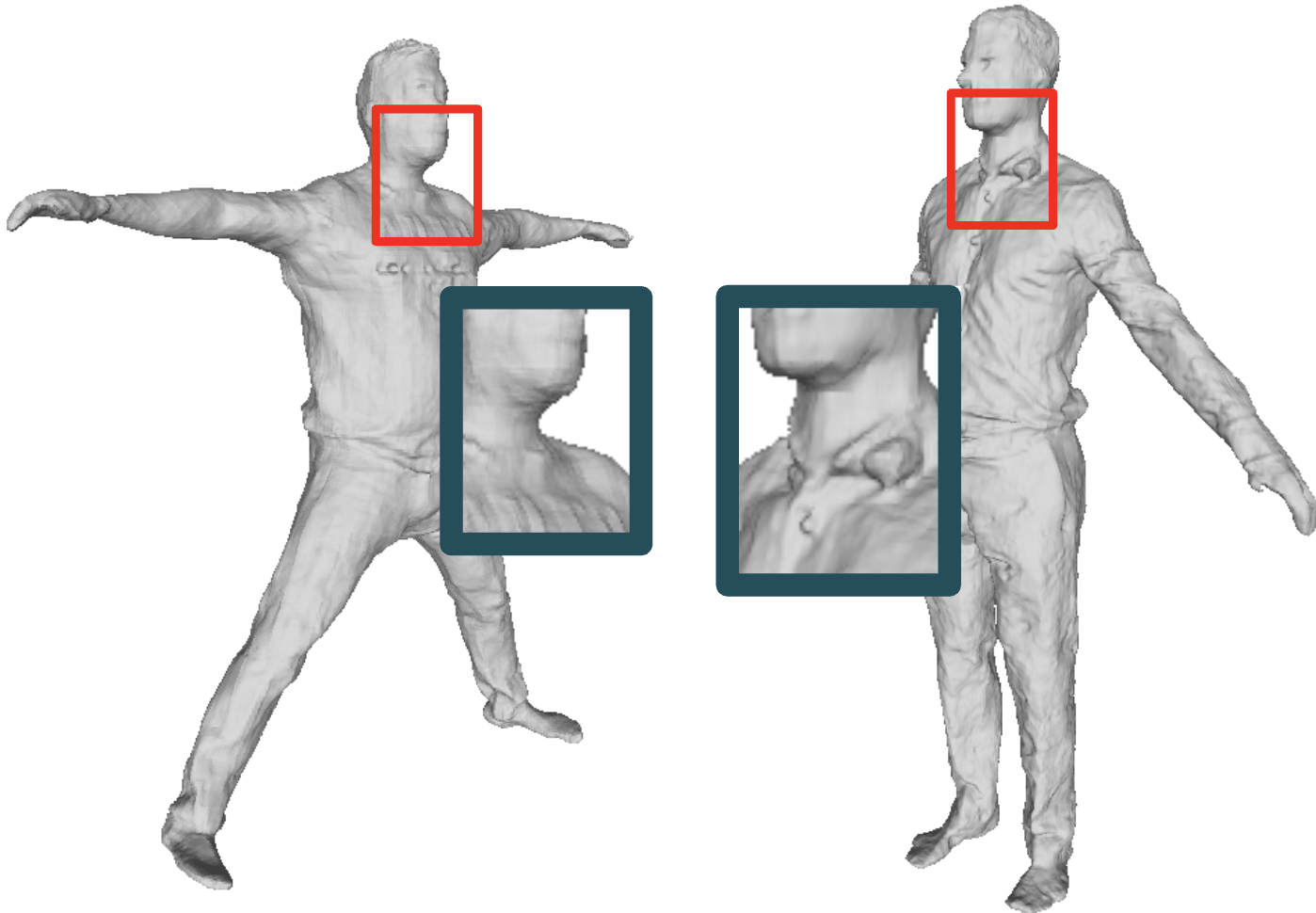}
    \caption{With smoothing term.}
    \label{fig:short-b}
  \end{subfigure}
  \caption{\textbf{Importance of Smooth Surface Regularization.}}
  \label{fig:withsmooth}
\end{figure}

We compare our approach against using numerical gradients and the Laplacian in the curvature loss~\cite{neuralangelo}, and against a hybrid positional encoding~\cite{shinobi}. The proposed SDF smoothing technique significantly boosts quality (see Table~\ref{tab:ablation}, and Figures~\ref{fig:comparison_methods}  and \ref{fig:withsmooth}). Even after training for several hours, the numerical gradients and the curvature loss fail to yield satisfactory results (Figure~\ref{fig:ablation-24h}).  Tweaking the weight of the Eikonal loss term is insufficient for good reconstruction quality (Figure~\ref{fig:comparison_methods}). Table~\ref{tab:ablation_speed} shows that our approach keeps the same speed as the base model and incurs less memory overhead than other approaches.

%% file: sections/05_limitations_future_work.tex
\section{Conclusions}
\label{sec:conclusions}

We proposed InstantGeoAvatar, a method that can model geometry and appearance of animatable human avatars from monocular videos in less than 10 minutes. Building on previous work, we notice that hash grid-based representations lack the implicit regularization that MLP-based architectures enjoy, and introduce a surface regularization term in the optimization that effectively enhances the learned representation without additional computational cost.